# A Robust Point Sets Matching Method[1]

Xiao Liu,　Congying Han,　Tiande Guo

School of Mathematical Science, University of Chinese Academy of Sciences, Beijing, 100049

**Abstract**
Point sets matching method is very important in computer vision, feature extraction, fingerprint matching, motion estimation and so on. This paper proposes a robust point sets matching method. We present an iterative algorithm that is robust to noise case. Firstly, we calculate all transformations between two points. Then similarity matrix are computed to measure the possibility that two transformation are both true. We iteratively update the matching score matrix by using the similarity matrix. By using matching algorithm on graph, we obtain the matching result. Experimental results obtained by our approach show robustness to outlier and jitter.

**Keywords**　 Point sets matching; Similarity matrix; Robust matching method; Point to point;

## 1. Introduction

The point sets matching problem is to find the optimum or a good suboptimal spatial mapping between the two point sets. In its general form, the point sets matching problem always consist of point sets with noise, outliers and geometric transformations. A general point sets matching algorithm needs to solve all these problems. It should be able to find the correspondences between two point sets, reject outliers and determine a transformation that can map one point set to the other.

Some popular class of methods solves the problem of affine point sets matching. Searching interest points and trying to search the affine invariant points in an affine Gaussian scale space was proposed in [1]. In [2], a developed algorithm with the thin-plate spline as the parameterization of the non-rigid spatial mapping and the soft assign for the correspondence was suggested. In [3], the authors proposed a novel approach for affine point pattern matching. When rotating one point sets to match the other point set, the value of a cost function which is invariant under special affine transformations would be calculated. When the value reached the minimum, the affine transformations were got. Using partial Hausdorff distance, a genetic algorithm (GA) based method to solve the incomplete unlabeled matching problem was presented in [4].

In [5], the authors proposed a sparse graphical model for solving the rigid point pattern matching.

Obviously, the point sets is larger, the matching time is much longer. To speed up the matching, it's better to match the subsets firstly to get the initial transformation parameters. Dror Aiger and Klara Kedem[6] proposed an asymptotically faster algorithm under rigid transformations, and provided new algorithms for homothetic

---

and similarity transformations.

The point sets matching problem arises in the domains of computer vision. A common application is image registration such as [7,8,9,10].

Another common application is super resolution which is the process of combining a sequence of low-resolution noisy blurred images to produce a higher resolution image or sequence. In [11], the authors proposed a method for matching feature points robustly across widely separated images.

In this paper, we propose a novel solution to the rigid point sets matching. Although we assume rigid motion, jitter and noise are allowed. This property is very important because noise and jitter are inevitable from the process of image acquisition and feature extraction. The rest of this paper is organized as follows. Section 2 introduces the assumption between point sets. Section 3 describes the main contribution of this paper, which is a robust matching algorithm to the rigid point sets. Simulations on synthetic data are presented in Section 4, and Section 5 concludes this paper.

## 2. Statement of the problem

Suppose two point sets A and B in two dimensions are given. That is $\{p_1, p_2 \ldots p_m\}$ and $B = \{q_1, q_2 \ldots q_n\}$, where the $p_i$ and $q_j$ are points in $R^2$. We want to find a global similarity transformation $T_{\theta, t_x, t_y}$, such that T(A) "matches" some subset of B, where matching will be made precise below. In the transformation $T_{\theta, t_x, t_y}$, $\theta$ is a rotation angle, $t_x$ and $t_y$ are the x and y translations, respectively. That is, for $(x, y) \in R^2$, we have

$$T\begin{pmatrix}x\\y\end{pmatrix} = \begin{pmatrix}t_x\\t_y\end{pmatrix} + \begin{pmatrix}\cos\theta & -\sin\theta\\ \sin\theta & \cos\theta\end{pmatrix}\begin{pmatrix}x\\y\end{pmatrix}. \qquad (1)$$

The first condition dictates that there exists a rigid transformation which should match most of points between A and B. In other word, only a few points in set are noise points which are random and disordered.

The second condition states the optimal transformation cause similarity between neighbors. Suppose $p_i$ and $p_j$ are neighbors in A, the transformation T matches $p_i$ and $p_j$ to $q_k$ and $q_l$, respectively. If T is optimal, $q_k$ and $q_l$ are like to be neighbors.

In this paper, we assume the point sets in $R^2$ is directed. The point feature, represented by feature location is the simplest form of feature. However, in many applications, such as motion estimation and fingerprint matching, the point features have both location and direction.

## 3. Robust directed point sets matching

We now briefly describe our proposed methodology.

To simplify the description, we use P and Q to represent an arbitrarily point in A and B, respectively. The coordinates of P and Q are $(P_x, P_y)$ and $(Q_x, Q_y)$, respectively. The direction of P and Q is $P_\theta$ and $Q_\theta$, respectively.

The transformation $T_{\theta,t_x,t_y}$ from P to Q, which means T(P)=Q, are firstly calculated.

Since points are directed, the rotate angle $\theta$ is $Q_\theta - P_\theta$. Then the transformation $T_{\theta,t_x,t_y}$ is got, and

$$\begin{cases} t_x = Q_x - (P_x \cos\theta - P_y \sin\theta) \\ t_y = Q_y - (P_x \sin\theta + P_y \cos\theta) \end{cases}. \tag{2}$$

The similarity between two transformations is considered. The correct matching should be similar to many other correct matching, while the incorrect matching not because of its randomness. If we randomly choose two matching and compute the similarity between them, the similarity between both true matching is significantly bigger than the similarity between both random matching.

Given two transformation $T(t_x, t_y, \theta)$ and $T'(t_x, t_y', \theta')$, we have the similarity as:

$$\text{similarity}(T, T') = \begin{cases} 0, & if\ |t_x - t_x'| > \alpha\ or\ |t_y - t_y'| > \beta\ or |\theta - \theta'| > \delta \\ 1 - \dfrac{\dfrac{|t_x-t_x'|}{\alpha} + \dfrac{|t_y-t_y'|}{\beta} + \dfrac{|\theta-\theta'|}{\delta}}{3}, & otherwise \end{cases}, \tag{3}$$

where $\alpha$ and $\beta$ are threshold of x and y, respectively, $\delta$ is threshold of angle.

Let us define $T_{ij}$ as the transformation $T(p_i) = q_j$. Take similarity between neighbor into consideration, we iteratively update matching score matrix using

$$W_{t+1}(i,j) = W_t(i,j) + \sum_{k \in SMN_{P,i}, l \in SMN_{Q,j}} W_t(k,l) * \text{similarity}(T_{ij}, T_{kl}), \tag{4}$$

where $W_t$ represents matching score matrix in the *t*th iteration, $SMN_{P,i}$ means the neighbor points of point i in A, $SMN_{Q,j}$ means the neighbor points of point j in B.

There are several ways to define the neighbor relationship in point sets. In this paper, we use the K nearest points in Euclidean distance to define the neighbor relationship. The magnitude of K will be discussed in experiments.

It is obvious that the matching score matrix always increase during the update. To transform the matching score matrix into the ideal score interval, matching score matrix will be linearly normalized after each iterative.

While the iterations stop, we model point sets matching as a weighted graph matching problem, where weights correspond to the matching score matrix. Then we can find a maximum score by Kuhn-Munkres algorithm.

## 4. Experiments

The synthetic point sets are generated as follows. First, generate a random point sets whose number, N, is given, named A. Then, another point sets B is generate from rigid rotation and translation on A. It is obvious that each point in A has a unique corresponding point in B. Randomly choose some point matching pairs from A to B, use the random noise points to replace them. The ratio of number of noise points to number of whole points is called outlier ratio. The rest matching pairs are added by a jitter value, which is random chose. The ratio of jitter value's range to whole point sets' range is called jitter ratio.

We use the average correct point pair ratio (ACPPR) to measure the robustness. The correct point pair ratio is defined as follows:

$$\text{Average Correct Point Pair Ratio} = \frac{\text{correct matching pairs' number by algorithm}}{\text{all the correct matching pairs' number}} \quad (5)$$

To reduce the uncertainty on initializing point sets, we repeated experiments and then calculated the average correct point pair ratio. In this paper, all ACCPR values are calculated from at least 20 simulation experiments.

Fig. 1 shows the ACPPR changing trend with the increasing K. The horizontal axis (K) is the number of neighbors to one point, dependent on different N which is the number of points of A. Table 1 lists the parameters used.

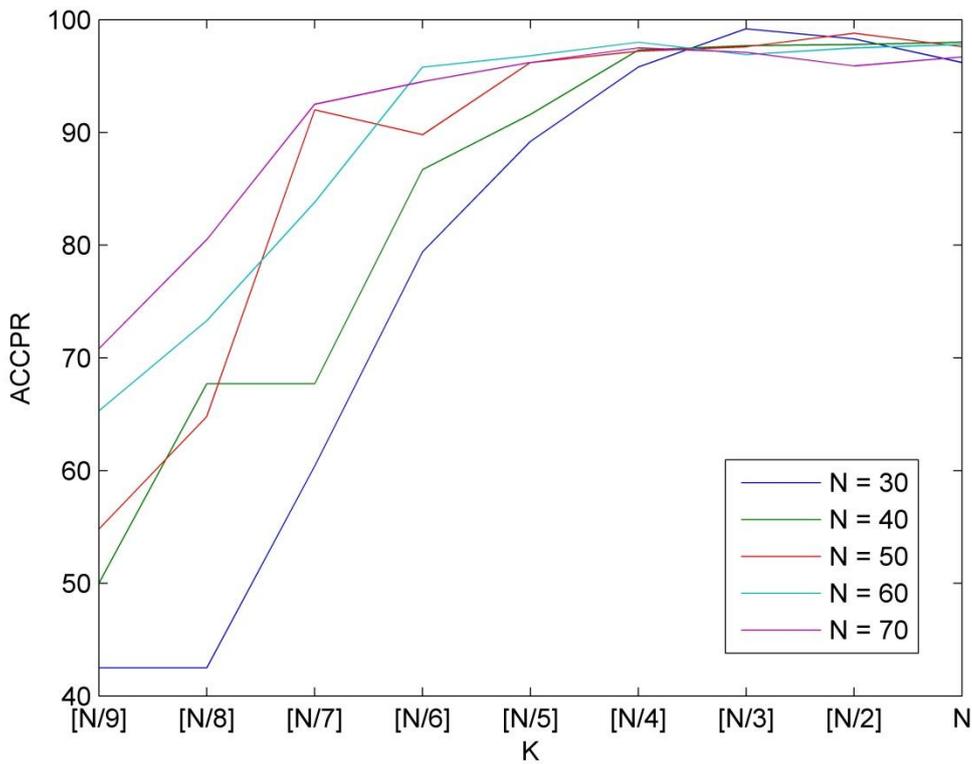

Fig.1 ACPPR comparison of different K.

In this experiment, outlier ratio is 20%, jitter ratio is 8%. And the $(\alpha, \beta, \delta)$, which are thresholds to measure the similarity between transformations, are $(10, 10, \pi/6)$.

From the analysis of Fig.1 we agree that ACPPR increases obviously and reaches to 95%, when K increases. An important result is that, when K is bigger than $[\frac{N}{4}]$,

ACPPR converges to a limit, no matter how many N is. It reminds us that if K is bigger enough to N, the magnitude of N is independent of ACPPR. The further discussion is shown below.

The ACPPR results with the different outlier ratio are compared in Fig.2. In this experiment, N is 50, jitter ratio is 8%, (α,β,δ) is (10, 10, π/6). The setting of K (6,12,25,50) represents $\left[\frac{N}{8}\right], \left[\frac{N}{4}\right], \left[\frac{N}{2}\right]$, N, respectively.

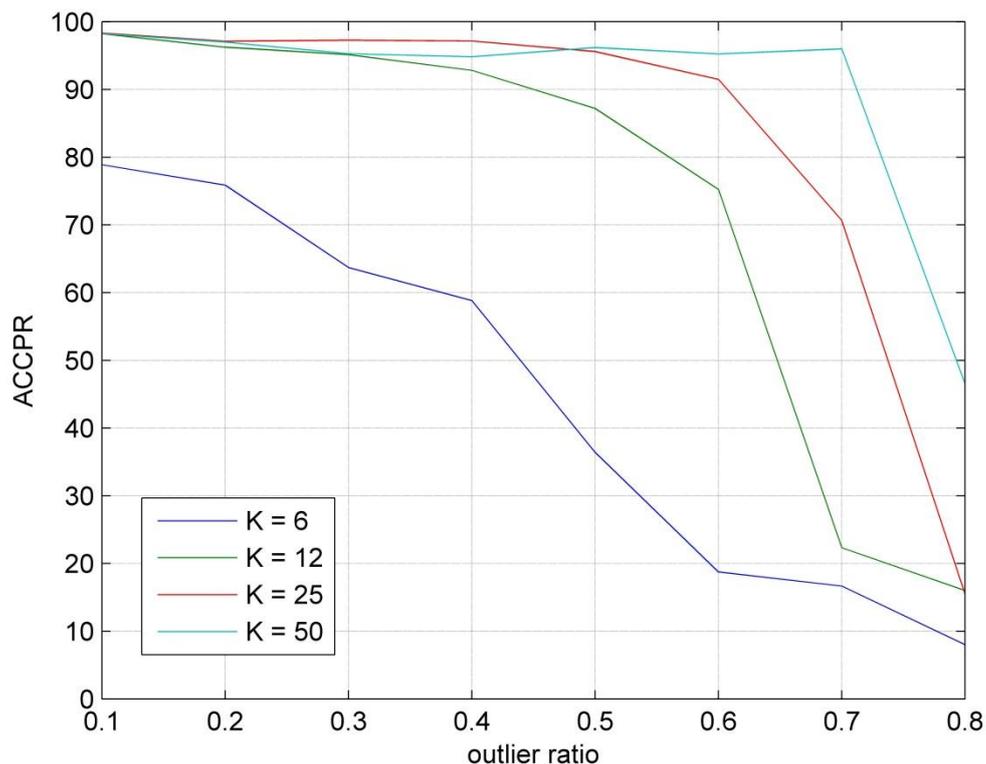

Fig.2 ACPPR comparison of different outlier ratio on different K.

While outlier ratio increases, noise points increase and true matching point pairs decrease. Take K equals 12 as an example, while outlier ratio increases from 10% to 50%, ACPPR slowly decreases from 98% to 87%. This phenomenon indicates that our method is robust when true pairs are more than noises. However, while outlier ratio increases from 50% to 80% , ACPPR rapidly decreases from 87% to 16%. This is an inevitable trend. When the outlier ratio is more than 50%, noise points are more than true matching point pairs, then the correct and meaningful information has been covered up by the wrong and meaningless information.

In Fig.3, we present the ACPPR with the different jitter ratio. In this experiment, N is 50, outlier ratio is 20%, (α,β,δ) is (10, 10, π/6). The setting of K (6,12,25,50) represents $\left[\frac{N}{8}\right], \left[\frac{N}{4}\right], \left[\frac{N}{2}\right]$, N, respectively.

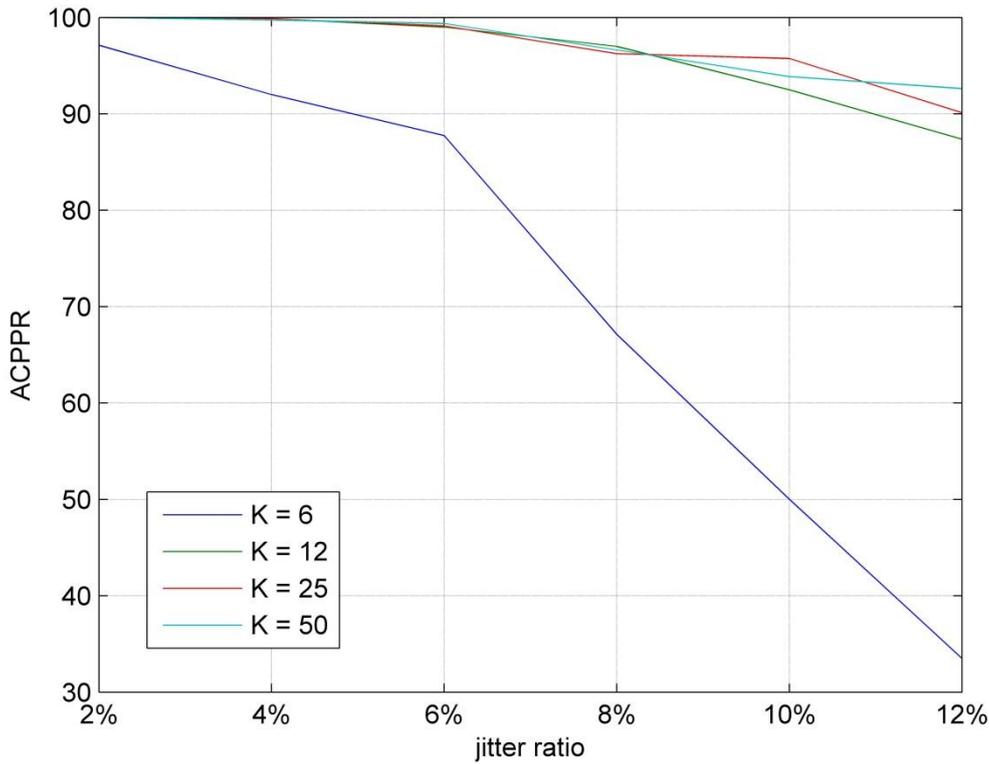

Fig.3 ACPPR comparison of different jitter ratio on different K.

When K is no less than $[\frac{N}{4}]$, ACPPR decreases slowly. While jitter ratio reaches 12%, ACPPR is more than 85%. This graph illustrates our method is robust on jitter ratio when K is not too much less than N.

We tested out method on the condition that both outlier ratio and jitter ratio are variable. The results are shown at table 1 as follows. In table 1 we show the performance of our algorithm for outlier ratio from 0% to 60%, different jitter ratio from 0% to 12%, (a),(b),(c),(d) are different of K, which are 6,12,25,50, respectively. All shown is Table 1, we set N = 50, (α, β, δ) = (10, 10, π/6).

Table 1 Comparisons of ACPPR for different K, outlier ratios and jitter ratios

(a) K=6

| ACPPR (K = 6) | | Jitter ratio (%) | | | | | | | Average (%) |
|---|---|---|---|---|---|---|---|---|---|
| | | 0 | 2 | 4 | 6 | 8 | 10 | 12 | |
| Outlier Ratio (%) | 0 | 100 | 100 | 90.4 | 90.4 | 74 | 50 | 36.8 | 77.4 |
| | 10 | 97.2 | 98.9 | 94.7 | 87.7 | 85.9 | 69.7 | 38.3 | 81.8 |
| | 20 | 92.3 | 93.9 | 91.8 | 88.9 | 72.4 | 55.9 | 28.4 | 74.8 |
| | 30 | 93.8 | 92.6 | 82.2 | 81.5 | 57.6 | 52.1 | 32.1 | 70.3 |
| | 40 | 81.3 | 77.1 | 77.5 | 75.6 | 60.7 | 33.5 | 22.7 | 61.2 |
| | 50 | 70.7 | 71.2 | 69.8 | 65.1 | 39.8 | 32.3 | 14.2 | 51.9 |
| | 60 | 60 | 55.4 | 49.2 | 48.8 | 21.2 | 11 | 4.4 | 35.7 |
| Average (%) | | 85 | 84.2 | 79.4 | 76.9 | 58.8 | 43.5 | 25.3 | 64.7 |

(b) K=12

| ACPPR (K = 12) | | Jitter ratio (%) | | | | | | | Average (%) |
|---|---|---|---|---|---|---|---|---|---|
| | | 0 | 2 | 4 | 6 | 8 | 10 | 12 | |
| Outlier Ratio (%) | 0 | 100 | 100 | 100 | 100 | 98.4 | 96 | 92 | 98.1 |
| | 10 | 100 | 100 | 99.5 | 98.4 | 98.5 | 93.7 | 88 | 96.9 |
| | 20 | 100 | 100 | 99.8 | 99 | 95.3 | 92.7 | 89.4 | 96.6 |
| | 30 | 99.9 | 100 | 99.7 | 98.7 | 92.4 | 88.5 | 66.7 | 92.3 |
| | 40 | 99.7 | 99.7 | 99.3 | 96.1 | 93.7 | 82.5 | 59.6 | 90.1 |
| | 50 | 98.9 | 98.1 | 99.2 | 95.4 | 87.7 | 74.4 | 53.6 | 86.8 |
| | 60 | 92 | 96.2 | 95.2 | 79.8 | 69.4 | 63.6 | 23.2 | 74.2 |
| Average (%) | | 98.6 | 99.1 | 99 | 95.3 | 90.8 | 84.5 | 67.5 | 90.7 |

(c) K=25

| ACPPR (K = 25) | | Jitter ratio (%) | | | | | | | Average (%) |
|---|---|---|---|---|---|---|---|---|---|
| | | 0 | 2 | 4 | 6 | 8 | 10 | 12 | |
| Outlier Ratio (%) | 0 | 100 | 100 | 100 | 100 | 98.4 | 97.6 | 92.8 | 98.4 |
| | 10 | 100 | 100 | 100 | 99.8 | 98.9 | 95.9 | 91.8 | 98.1 |
| | 20 | 100 | 100 | 100 | 99.4 | 98.6 | 95.8 | 93.2 | 98.1 |
| | 30 | 100 | 100 | 100 | 99.4 | 97.1 | 94.6 | 88.3 | 97.1 |
| | 40 | 100 | 99.7 | 99.5 | 98.8 | 96.8 | 92 | 88.3 | 96.4 |
| | 50 | 100 | 99.4 | 98.9 | 98.2 | 96.6 | 94.4 | 84.2 | 96 |
| | 60 | 100 | 99.4 | 98.2 | 98 | 90.4 | 75.6 | 70.6 | 90.3 |
| Average (%) | | 100 | 99.8 | 99.5 | 99.1 | 96.7 | 92.3 | 87 | 96.3 |

(d) K=50

| ACPPR (K = 50) | | Jitter ratio (%) | | | | | | | Average (%) |
|---|---|---|---|---|---|---|---|---|---|
| | | 0 | 2 | 4 | 6 | 8 | 10 | 12 | |
| Outlier Ratio (%) | 0 | 100 | 100 | 100 | 100 | 99.2 | 98.4 | 98.4 | 99.4 |
| | 10 | 100 | 100 | 100 | 99.1 | 97.9 | 96.8 | 92.2 | 98 |
| | 20 | 100 | 100 | 100 | 99.1 | 98.2 | 96.6 | 93.9 | 98.3 |
| | 30 | 100 | 100 | 99.9 | 98.6 | 96.9 | 93.8 | 89.9 | 97 |
| | 40 | 100 | 100 | 99.7 | 98.8 | 98.1 | 94.1 | 88.8 | 97.1 |
| | 50 | 100 | 100 | 98.7 | 98.2 | 97.1 | 93.1 | 89.8 | 96.7 |
| | 60 | 99.6 | 99.8 | 99.6 | 98.6 | 96.8 | 89.4 | 87.4 | 95.9 |
| Average (%) | | 99.9 | 100 | 99.7 | 98.9 | 97.7 | 94.6 | 91.5 | 97.5 |

When ACPPR in one sub table, take table 1(c) as an example, are compared, it is easy to find that the ACPPR decrease as the jitter ratio or outlier ratio increases. While outlier ratio increases quickly from 0% to 50%, ACPPR decreases slowly and is 70.6% even in the worst situation (outlier ratio = 60%, jitter ratio = 12%). While jitter ratio increases from 0% to 12%, ACPPR decreases from 100% to 87% in average ACCPR of all outlier ratios. We compare average ACPPR among different sub tables. Our method performs badly when K is small as 6. This is because the size of K decides the number of neighbor points. The more neighbor points, the more information is got. Small K results in lack of robustness.

## 5. Conclusion

In this paper, we have proposed a novel method for point sets matching. While our method resembles other point sets matching methods in calculating transformation, our method is novel in that it uses the similarity between transformations and the influence of neighbors. More specifically, our method does not depend on only point location, but also the point direction. An important implication of this property is that we can expect our method to work on some special situation, such as motion estimation in super resolution.

Besides, we believe the use of direction and similarity of transformation can better matching result.

## Acknowledgments

This work was funded by the Chinese National Natural Science Foundation （11331012, 71271204 and 11101420）.